# Evolutionary Algorithms: Concepts, Designs, and Applications in Bioinformatics

## Evolutionary Algorithms for Bioinformatics

Ka-Chun Wong
*Department of Computer Science, University of Toronto, Ontario, Canada*

## INTRODUCTION

Since genetic algorithm was proposed by John Holland (Holland J. H., 1975) in the early 1970s, the study of evolutionary algorithm has emerged as a popular research field (Civicioglu & Besdok, 2013). Researchers from various scientific and engineering disciplines have been digging into this field, exploring the unique power of evolutionary algorithms (Hadka & Reed, 2013). Many applications have been successfully proposed in the past twenty years. For example, mechanical design (Lampinen & Zelinka, 1999), electromagnetic optimization (Rahmat-Samii & Michielssen, 1999), environmental protection (Bertini, Felice, Moretti, & Pizzuti, 2010), finance (Larkin & Ryan, 2010), musical orchestration (Esling, Carpentier, & Agon, 2010), pipe routing (Furuholmen, Glette, Hovin, & Torresen, 2010), and nuclear reactor core design (Sacco, Henderson, Rios-Coelho, Ali, & Pereira, 2009). In particular, its function optimization capability was highlighted (Goldberg & Richardson, 1987) because of its high adaptability to different function landscapes, to which we cannot apply traditional optimization techniques (Wong, Leung, & Wong, 2009).

## BACKGROUND

Evolutionary algorithms draw inspiration from nature. An evolutionary algorithm starts with a randomly initialized population. The population then evolves across several generations. In each generation, fit individuals are selected to become parent individuals. They cross-over with each other to generate new individuals, which are subsequently called offspring individuals. Randomly selected offspring individuals then undergo certain mutations. After that, the algorithm selects the optimal individuals for survival to the next generation according to the survival selection scheme designed in advance. For instance, if the algorithm is overlapping (De Jong, 2006), then both parent and offspring populations will participate in the survival selection. Otherwise, only the offspring population will participate in the survival selection. The selected individuals then survive to the next generation. Such a procedure is repeated again and again until a certain termination condition is met (Wong, Leung, & Wong, 2010). Figure 1 outlines a typical evolutionary algorithm

```
Algorithm 1 A Typical Evolutionary Algorithm
    Choose suitable representation methods;

    P(t): Parent Population at time t
    O(t): Offspring Population at time t

    t ← 0;
    Initialize P(t);
    while not termination condition do
        temp = Parent Selection from P(t);
        O(t + 1) = Crossover in temp;
        O(t + 1) = Mutate O(t + 1);
        if overlapping then
            P(t + 1) = Survival Selection from O(t + 1) ∪ P(t) ;
        else
            P(t + 1) = Survival Selection from O(t + 1) ;
        end if
        t ← t + 1;
    end while

    Good individuals can then be found in P(t);
```

*Figure 1. Major components of a typical evolutionary algorithm*

In this book chapter, we follow the unified approach proposed by De Jong (De Jong, 2006). The design of evolutionary algorithm can be divided into several components: representation, parent selection, crossover operators, mutation operators, survival selection, and termination condition. Details can be found in the following sections.

- **Representation:** It involves genotype representation and genotype-phenotype mapping. (De Jong, 2006). For instance, we may represent an integer (phenotype) as a binary array (genotype): '19' as '10011' and '106' as '1101010'. If we mutate the first bit, then we will get '3' (00011) and '42' (0101010). For those examples, even we have mutated one bit in the genotype, the phenotype may vary very much. Thus we can see that there are a lot of considerations in the mapping.

- **Parent Selection:** It aims at selecting good parent individuals for crossovers, where the goodness of a parent individual is quantified by its fitness. Thus most parent selection schemes focus on giving more opportunities to the fitter parent individuals than the other individuals and vice versa such that "good" offspring individuals are likely to be generated.

- **Crossover Operators:** It resembles the reproduction mechanism in nature. Thus they, with mutation operators, are collectively called reproductive operators. In general, a crossover operator combines two individuals to form a new individual. It tries to split an individual into parts and then assemble those parts into a new individual.

- **Mutation Operators:** It simulates the mutation mechanism in which some parts of a genome undergoes random changes in nature. Thus, as a typical modeling practice, a mutation operator changes parts of the genome of an individual. On the other hand, mutations can be thought as an exploration mechanism to balance the exploitation power of crossover operators.

- **Survival Selection:** It aims at selecting a subset of good individuals from a set of individuals, where the goodness of individual is proportional to its fitness in most cases. Thus survival selection mechanism is somehow similar to parent selection mechanism. In a typical framework like 'EC4' (De Jong, 2006), most parent selection mechanisms can be re-applied in survival selection.

- **Termination Condition:** It refers to the condition at which an evolutionary algorithm should end.

# EVOLUTIONARY ALGORITHMS: CONCEPTS AND DESIGNS

## Representation

Representation involves genotype representation and genotype-phenotype mapping. In general, designers try to keep genotype representation as compact as possible while keeping it as close to the corresponding phenotype representations as possible such that measurement metrics, say distance, in the genotype space can be mapped to those in phenotype space without the loss of semantic information.

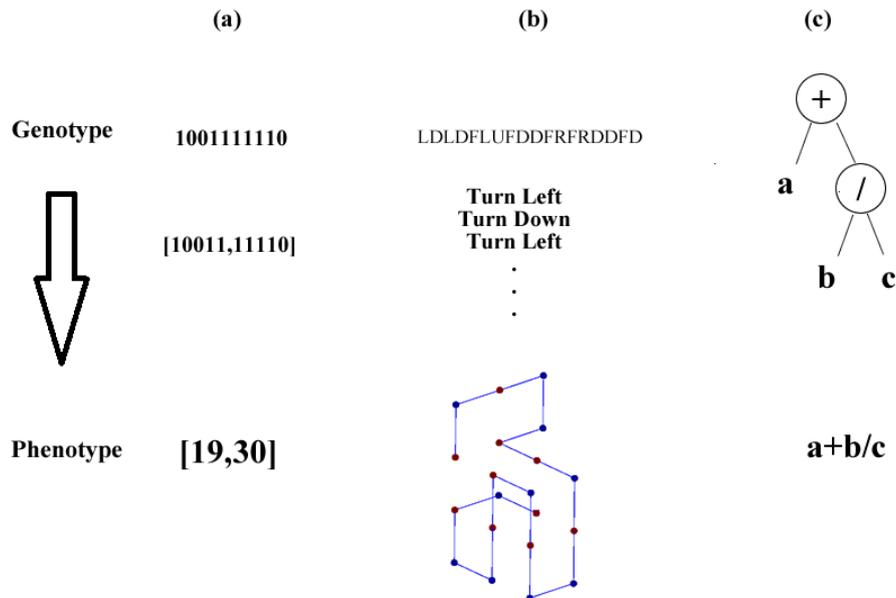

*Figure 2. Some representations of evolutionary algorithms: (a) Integer representation (b) Protein structure representation on a lattice model (c) Tree representation for a mathematical expression*

In general, there are many types of representations that an evolutionary algorithm can adopt. For example, fixed-length linear structures, variable-length linear structures, and tree structures…… Figure 2 depicts three examples. Figure 2a is a vector of integers. We can observe that its genotype is a binary array with length equal to 10. To map it into the phenotype space, the first 5 binary digits (10011) are mapped to the first element (19) of the vector whereas the remaining 5 binary digits (11110) are mapped to the second element (30) of the vector. Figure 2b is the relative encoding representation of a protein on the HP lattice model (Krasnogor, Hart, Smith, & Pelta, 1999). Its genotype is an array of moves and its length is set to the amino acid sequence length of the protein. The array of moves encodes the relative positions of amino acids from their predecessor amino acids. Thus we need to follow the move sequence to compute the 3D structure of the protein (phenotype) for further evaluations. Figure 2c is the tree representation of a mathematical expression. Obviously, such tree structure is a variable length structure, which has the flexibility in design. If the expression is short, it can be shrunk during the evolution. If the expression is long, it can also be expanded during the evolution. Thus we can observe that the structure has an advantage over the previous representations. Nevertheless, there is no free lunch. It imposes several implementation difficulties to translate it into phenotypes.

## Parent Selection

Parent selection aims at selecting "good" parent individuals for crossover, where the goodness of a parent individual is positively proportional to its fitness for most cases. Thus most parent selection schemes focus on giving more opportunities to the fitter parent individuals than the other individuals and vice versa. The typical methods are listed as follows:

- **Fitness Proportional Selection:** The scheme is sometimes called roulette wheel selection. In the scheme, the fitness values of all individuals are summed. Once summed, the fitness of each individual is divided by the sum. The ratio then becomes the probability for each individual to be selected.

- **Rank Proportional Selection:** Individuals with high ranks are given more chances to be selected. Unlike fitness proportional scheme, the rank proportional scheme does not depend on the actual fitness values of the individuals. It is a double-edged sword. On the positive side, it can help us prevent the domination of very high fitness values. On the negative side, it imposes additional computational costs for ranking.

- **Uniform Deterministic Selection:** The scheme is the simplest among the other schemes. All individuals are selected, resulting in uniform selection.

- **Uniform Stochastic Selection:** The scheme is the probabilistic version of uniform deterministic selection. All individuals are given equal chances (equal probabilities) to be selected.

- **Binary Tournament:** Actually, there are other tournament selection schemes proposed in the past literature. In this book chapter, the most basic one, binary tournament, is selected and described. In each binary tournament, two individuals are randomly selected and competed with each other by fitness. The winner is then selected. Such a procedure is repeated until all vacancies are filled.

- **Truncation:** The top individuals are selected deterministically when there is a vacancy for selection. In other words, the bottom individuals are never selected. For example, if there are 100 individuals and 50 slots are available, then the top 50 fittest individuals will be selected.

## Crossover Operators

Crossover operators resemble the reproduction mechanism in nature. Thus they, with mutation operators, are collectively called reproductive operators. In general, a crossover operator combines two individuals to form a new individual. It tries to partition an individual into parts and then assemble the parts of two individuals into a new individual. The partitioning is not a trivial task. It depends on the representation adopted. Thus it is not hard to imagine that crossover operators are representation-dependent. Nevertheless, without loss of generality, a list of classic crossover operators is listed as follows:

- **One Point Crossover:** One point crossover is a commonly used crossover operator because of its simplicity. Given two individuals, it randomly chooses a cut point in their genomes. Then it swaps the parts after (or before) the cut point between the two genomes.

- **Two Points Crossover:** Two points crossover is another commonly used crossover operator because people argue that one point crossover has a positional bias toward the terminal positions. For instance, when making a one point crossover, the rightmost (or leftmost) part is always swapped. Thus people propose two point crossovers to avoid the positional bias.

- **Uniform Crossover:** Uniform crossover is a general one. Each gene is given an equal probability to be swapped.

- **Blend Crossover:** Blend crossover is commonly used in real number optimization. Instead of swapping genes, it tries to blend two genes together by arithmetic averaging to obtain the intermediate values. For instance, if we are going to make a crossover between two vectors [1 2 3] and [4 5 6], then the blended vector will be [2.5 3.5 4.5]. Weights can be applied here.

## Mutation Operators

Mutation operators resemble the mutation mechanism in which some parts of genome undergo random changes in nature. Thus, as a typical modeling, a mutation operator changes parts of the genome of an individual probabilistically. Similar to crossover operators, mutation operators are representation-dependent. Nevertheless, without loss of generality, a list of commonly used mutation operators is shown below:

- **Bitflip Mutation:** It is commonly used in binary genomes. Specified by a pre-defined probability, each bit in a binary genome is probabilistically inverted.

- **Random Mutation:** Random mutation is generalized from bitflip mutation. It can be applied in many genomes. Specified by a pre-defined probability, each part in a genome is probabilistically changed to a random value within domain bounds.

- **Delta Mutation:** Delta mutation is commonly used in real number genomes. Specified by a pre-defined probability, each real number in a real number genome is probabilistically incremented/decremented by a certain step size (called delta), where the step size is pre-specified. Nonetheless, it is straightforward to make the step size adaptive, similar to the trial vector generations in differential evolution (Storn & Price, 1997).

- **Gaussian Mutation:** Gaussian mutation is also commonly used in real number genomes. Similar to delta mutation, each real number in a real number genome is probabilistically increased / decreased by a step size. The difference is that the step size is a Gaussian random number. (De Jong, 2006).

## Survival Selection

Survival selection aims at selecting a subset of good individuals from a population, where the goodness of individual is proportional to its fitness for most cases. Thus survival selection mechanism is somehow similar to parent selection mechanism. In a typical framework like EC4 (De Jong, 2006), most parent selection mechanisms can be re-applied in survival selection. For example, the fitness proportional selection can be applied as survival selection.

## Termination Condition

Termination condition refers to the condition at which an evolutionary algorithm should end. For historical reasons, the number of generations is often adopted as the termination measurement: an evolutionary algorithm terminates when a certain number of generations has been reached (e.g. 1000 generations). Nonetheless, it has been pointed out that fitness function evaluations are computationally expensive in certain domains. Thus the number of fitness function evaluations is also adopted in some problems. If computing resources are limited, CPU time is also adopted. Nonetheless, convergence is not guaranteed. Thus people have calculated the fitness improvement of each generation as another condition for termination.

## Examples

**Genetic Algorithm:** Genetic algorithm is the most classic evolutionary algorithm. It draws inspiration from the Darwin's Evolution Theory. The difference between genetic algorithm and evolutionary algorithm becomes blurred nowadays. The words 'genetic algorithm' and 'evolutionary algorithm' are sometimes interchanged in use. To clearly explain the working mechanism of a genetic algorithm, we chose the canonical genetic algorithm (Whitley, 1994) as a representative example.

In the canonical genetic algorithm, each individual has a fixed-length binary array as its genotype. Then the fitness of each individual is divided by the average fitness to calculate the normalized probability to be selected. The algorithm then adopts them to select parents for one point crossover to produce offspring individuals, which subsequently undergo mutations. The offspring individuals become the population in the next generation and so forth.

**Genetic Programming:** Genetic programming is indeed a special type of genetic algorithm. The difference lies in their representations. Genetic programming adopts trees as genotypes to represent programs or expressions. (Figure 2 depicts an example). The typical selection schemes of evolutionary algorithms can still be used as parent selection and survival selection in genetic programming. The distinct features of genetic programming are their crossover and mutation operators. For instance, swapping sub-trees between two trees and random generation of sub-trees. A list of common crossover and mutation operators for genetic programming is tabulated in Table 1.

*Table 1. A list of crossover and mutation operators for genetic programming* (Banzhaf, Nordin, Keller, & Francone, 1998)

|  | **Description** |
|---|---|
| Crossover | • Subtree Exchange Crossover: exchange subtrees between individuals<br>• Self Crossover: exchange subtrees within an individual<br>• Module Crossover: exchange modules between individuals<br>• SCPC: exchange subtrees if coordinates match exactly<br>• WCPC: exchange subtrees if coordinates match approximately |
| Mutation | • Point Mutation: change the value of a node<br>• Permutation: change the argument order of a node<br>• Hoist: use a subtree to become a new individual<br>• Expansion Mutation: exchange a subtree against a terminal node<br>• Collapse Subtree Mutation: exchange a terminal node against a subtree<br>• Subtree Mutation: replace a subtree by another subtree<br>• Gene Duplication: replace a subtree by a terminal |

**Differential Evolution:** Differential Evolution was first proposed by Price and Storn in the 1990s (Storn & Price, 1997). It demonstrated great potential for real function optimization in the subsequent contests (Price, 1997). Without loss of generality, a typical strategy of differential evolution (DE/rand/1) (Feoktistov, 2006) is shown in Figure 3.

```
Algorithm 2 Differential Evolution
  P_t: Population at time t
  TP: Transient population

  t ← 0;
  Initialize P_t;
  Evaluate P_t;
  while not termination condition do
      TP ← ∅;
      for ∀indiv_i ∈ P_t do
          Offspring ← TRIALVECTORGENERATION(indiv_i);
          Evaluate Offspring;
          if Offspring is fitter than indiv_i then
              Put Offspring into TP;
          else
              Put Parent into TP;
          end if
      end for
      t = t + 1;
      P_t ← TP;
  end while
```

*Figure 3. Outline of differential evolution (DE/rand/1)*

For each individual in a generation, the algorithm randomly selects three individuals to form a trial vector. One individual forms a base vector, whereas the value difference between the other two individuals forms a difference vector. The sum of those two vectors forms a trial vector, which recombines with the individual to form an offspring. Replacing the typical crossover and mutation operation by this trial vector generation, manual parameter tuning of crossover and mutation is no longer needed. It can provide differential evolution a self-organizing ability and high adaptability for choosing suitable step sizes which demonstrated its potential for continuous optimization in the past contests. A self-organizing ability is granted for moving toward the optima. A high adaptability is achieved for optimizing different landscapes (Feoktistov, 2006). With such self-adaptability, differential evolution is considered as one of the most powerful evolutionary algorithms for real function optimization. For example, mechanical engineering design (Lampinen & Zelinka, 1999) and nuclear reactor core design (Sacco, Henderson, Rios-Coelho, Ali, & Pereira, 2009).

**Evolution Strategy:** Evolution Strategy was proposed in 1968 (Beyer & Schwefel, 2002). It is even older than genetic algorithm. Schwefel and Klockgether originally used evolution strategy as a heuristic to perform several experimental optimizations in air flow. They found that evolution strategy was better than other discrete gradient-oriented strategy, which raised people's interests in evolution strategy. Comparing to the previous evolutionary algorithms, evolution strategy draws less inspiration from nature. Instead, it was artificially created as a numerical tool for optimization. Thus the structure of evolution strategy is quite different from the other evolutionary algorithms. For example, evolution strategy scholars call the mutation step size and probability as endogenous parameters encoded in the genome of an individual. Thus, besides the gene values, a genome is also composed of the parameter settings which control the convergence progress of the whole algorithm. The notation of evolution strategy is quite interesting. ($\mu / \rho^+, \lambda$) - ES denotes an evolution strategy where $\mu$ denotes parent population size; $\rho$ denotes breeding size; ($\mu / \rho + \lambda$) - ES denotes the algorithm is overlapping; ($\mu / \rho, \lambda$) denotes the algorithm is not overlapping; $\lambda$ denotes the offspring population size.

**Swarm Intelligence:** Ant Colony Optimization (Dorigo & Gambardella, 1997), Particle Swarm Optimization (Poli, Kennedy, & Blackwell, 2007), and Bee Colony Optimization (Karaboga, Akay, & Ozturk, 2007)……etc are collectively known as Swarm Intelligence. Swarm intelligence is a special class of evolutionary algorithm. It does not involve any selection (i.e. birth and death). Instead, it maintains a fixed-size population of individuals for search across generations. After each generation, the individuals report their findings which are recorded and used to adjust the search strategy in the next generation. Some of the algorithms were originally designed for shortest path finding. Nevertheless, people have further generalized them for other applications. For instance, Bi-Criterion Opitmization (Iredi, Merkle, & Middendorf, 2000), Load Balancing in Telecommunication Network (Schoonderwoerd, Bruten, Holland, & Rothkrantz, 1996), Protein Folding Problem (Shmygelska & Hoos, 2005), and Power System (del Valle, Venayagamoorthy, Mohagheghi, Hernandez, & Harley, 2008).

**Multimodel Optimization:** Real world problems always have different multiple solutions. For instance, optical engineers need to tune the recording parameters to get as many optimal solutions as possible for multiple trials in the varied-line-spacing holographic grating design problem because the design constraints are too difficult to be expressed and solved in mathematical forms. Unfortunately, most traditional optimization techniques focus on solving for a single optimal solution. They need to be applied several times; yet all solutions are not guaranteed to be found. Thus the multimodal optimization problem was proposed. In that problem, we are interested in not only a single optimal point, but also the others. Given an objective function, an algorithm is expected to find all optimal points in a single run. With strong parallel search capability, evolutionary algorithms are shown to be particularly effective in solving this type of problem. Although the objective is clear, it is not easy to be satisfied in practice because some problems may have too many optima to be located. Nonetheless, it is still of great interest to researchers how these problems are going to be solved because the algorithms for multimodal optimization usually not only locate multiple optima in a single run, but also preserve their population diversity throughout a run, resulting in their global optimization ability on multimodal functions. The work by De Jong (De Jong, 2006) is one of the first known attempts to solve the multimodal optimization problem by an evolutionary algorithm. He introduced the crowding technique to increase the chance of locating multiple optima. In the crowding technique, an offspring replaces the parent which is most similar to the offspring itself. Such a strategy can preserve the diversity and maintain different niches in a run. Twelve years later, Goldberg and Richardson (Goldberg & Richardson, 1987) proposed a fitness-sharing niching technique as a diversity preserving strategy to solve the multimodal optimization problem. They proposed a shared fitness function, instead of an absolute fitness function, to evaluate the fitness of a individual in order to favor the growth of the individuals which are distinct from the others. With this technique, a population can be prevented from the domination of a particular type of individuals. Species conserving genetic algorithm (SCGA) (Wong, Leung, & Wong, An evolutionary algorithm with species-specific explosion for multimodal optimization, 2009) is another technique for evolving parallel subpopulations. Before crossovers in each generation, the algorithm selects a set of species seeds which can bypass the subsequent procedures to the next generation. Since then, many researchers have been exploring different ways to deal with the problem. Notably, SCGA was claimed that the technique was considered as an effective and efficient method for inducing niching behavior into GAs. However, in our experiments, we find that the performance of the technique still has space for improvement. It always suffers from genetic drifts though each species is conserved with one individual. The results of the comparison test conducted by Singh et al. (Singh & Deb, 2006) also reveals that the species conserving technique performs the worst among the algorithms tested. As a result, Wong et al. have proposed a novel algorithm to remedy the species conserving technique. The novel algorithm is called Evolutionary Algorithm with Species-specific Explosion (EASE) for multimodal optimization (Wong, Leung, & Wong, An evolutionary algorithm with species-specific explosion for multimodal optimization, 2009). EASE is built on the Species Conserving Genetic Algorithm (SCGA), and the design is improved in several ways. In particular, it not only identifies species seeds, but also exploits the species seeds to create multiple mutated copies in order to further converge to the respective optimum for each species. Evolutionary Algorithm with Species-specific Explosion (EASE) is an evolutionary algorithm which identifies and exploits species seeds to locate global and local optima. There are two stages in the algorithm: Exploration Stage and Species-specific Stage. The exploration stage targets for roughly locating all global and local optima. It not only undergoes normal genetic operations: selection and crossover, but also involves the addition of randomly generated individuals for preserving the diversity. On the other hand, the species-specific stage targets for gently locating the optimum for each species. Species-specific genetic operations are applied. Only the individuals within the same species are allowed to perform selection and crossover to each other. No inter-species selection and crossover are allowed. Such a strategy is to provide more chances for each species to converge to its respective optimum, with the trade-off that diversity is no longer preserved. To have a better global picture for locating optima, EASE starts with the exploration stage. It will switch to the species-specific stage only after the stage switching condition is satisfied. No matter in which stage, a local operation called Species-specific Explosion is always executed so as to help species to climb and converge to its corresponding optimum. Briefly, in SCGA, Li et al. proposed conserving one individual for each species. However, just one individual for each species is not enough for the algorithm to well-conserve and nurture the species. In a run of SCGA, it is often the case that the algorithm does conserve species with low fitness values, but they are present in a small proportion. Once they form new offspring, their offspring are often removed quickly in subsequent generations due to their low fitness values. Thus most individuals are always of the species with high fitness values. In atypical run of SCGA, we can observe that the individuals gradually converge to the three optima fitness-proportionally. Though different species are preserved with an individual as the species seed, it cannot converge to some of the low-fitness local optima. Merely SCGA itself actually cannot provide enough indiscriminate condition for species to nurture, evolve, and converge to its respective optimum in each run. Hence EASE incorporates a local operation called Species-specific Explosion to nurture species and remedy their convergences. Species-specific explosion is the local operation in which we create multiple copies for each species seed and mutate them. In summary, EASE is divided into two stages: Exploration Stage and Species-specific Stage. EASE starts with the exploration stage. Once the stage switching condition is satisfied, it will be changed to species-specific stage. Throughout the two stages, a local operation: Species-specific Explosion is applied so as to help each species to converge to its respective optimum.

Though different methods were proposed in the past, they were all based on the same fundamental idea: it is to strike an optimal balance between convergence and population diversity in order to locate optima.

**Others:** Other evolutionary computation methods have been proposed; for instance, Cuckoo-search (Civicioglu & Besdok, 2013), Lévy flight (Vuswabatgab, Afanasyer, Buldyrev, Murphy, Prince, & Stanley, 1996), Bacterial Colony Optimization (Niu & Wang, 2012), and Intelligent Water Drops algorithm (Shah-Hosseini, 2009).

## APPLICATIONS TO BIOINFORMATICS

### An Overview of Bioinformatics

Since the 1990s, the whole genomes of a large number of species have been sequenced by their corresponding genome sequencing projects. In 1995, the first free-living organism *Haemophilus influenzae* was sequenced by the Institute for Genomic Research (Fleischmann, et al., 1995). In 1996, the first eukaryotic genome (*Saccharomyces cerevisiase*) was completely sequenced (Goffeau, et al., 1996). In 2000, the first plant genome *Arabidopsis thaliana*, was also sequenced by Arabidopsis Genome Initiative (Initiative, 2000). In 2004, the Human Genome Project (HGP) announced its completion (Consortium I. H., 2004). Following the HGP, the Encyclopedia of DNA Elements (ENCODE) project was started, revealing massive functional putative elements on the human genome in 2011 (Consortium E. , 2012). The drastically decreasing cost of sequencing also enables the 1000 Genomes Project to be carried out, resulting in an integrated map of genetic variation from 1,092 human genomes published in 2012 (Abecasis, Auton, Brooks, DePristo, & Durbin, 2012). Nonetheless, the massive genomic data generated by those projects impose an unforeseen challenge for large-scale data analysis at the scale of gigabytes or even terabytes (Wong, Peng, Li, & Chan, 2014).

In particular, computational methods are essential in analyzing the massive genomic data (Wong, Li, Peng, & Zhang, 2015). They are collectively known as bioinformatics or computational biology. For instance, motif discovery (GuhaThakurta, 2006) helps us distinguish real signal subsequence patterns from background sequences. Multiple sequence alignment (Altschul, Gish, Miller, Myers, & Lipman, 1990) can be used to analyze the similarities between multiple sequences. Protein structure prediction (McGuffin, Bryson, & Jones, 2000) can be applied to predict the 3D tertiary structure from an amino acid sequence. Gene network inference (D'Haeseleer, Liang, & Somogyi, 2000) are the statistical methods to infer gene networks from correlated data (e.g. microarray data). Promoter prediction (Abeel, Van de Peer, & Saeys, 2009) help us annotate the promoter regions on a genome. Phylogenetic tree inference (Ronquist & Huelsenbeck, 2003) can be applied to study the hierarchical evolution relationship between different species. Drug scheduling (Liang, Leung, & Mok, 2008) can help solve the clinical scheduling problems in an effective manner. Although the precisions of those computational methods are usually lower than the existing biotechnology, they can still serve as useful preprocessing tools to significantly narrow search spaces (Wong & Zhang, 2014). Thus prioritized candidates can be selected for further validation by wet-lab experiments, saving manual time and funding (Wong, Chan, Peng, Li, & Zhang, 2013).

**Evolutionary Algorithms for Protein Structure Prediction**

A polypeptide is a chain of amino acid residues. Once folded into its native state, it is called protein. Proteins play vital roles in living organisms. They perform different tasks to maintain a body's life. For instance, material transportations across cells and catalyzing metabolic reactions and body defenses against viruses. Nevertheless, functions of proteins substantially depend on their structural features. In other words, researchers need to know a protein's native structure before its function can be completely deduced. It gives rises to the protein structure prediction problem.

The protein structure prediction problem is often referred as the ``holy grail" of biology. In particular, Anfinsen's dogma (Anfinsen, 1973) and Levinthal's paradox (Levinthal, 1968) are the central rules in this problem. Anfinsen's dogma postulates that a protein's native structure (tertiary structure) only depends on its amino acid residue sequence (primary structure). On the other hand, Levinthal's paradox postulates that it is too time-consuming for a protein to randomly sample all the feasible confirmation regions for its native structure. On the other hand, the proteins in nature can still spontaneously fold into its native structures in about several milliseconds.

Based on the above ideas, researchers have explored the problem throughout several years. In particular, the protein structural design and sequence degeneracy have been studied by Li et. al. (Li, Helling, Tang, & Wingreen, 1996). The computational complexity has also been examined (Aluru, 2005).

Numerous prediction approaches have been proposed. In general, they can be classified into two categories, depending on whether any prior knowledge other than sequence data has been incorporated (Baker & Sali, 2001). This book chapter focuses on *de novo* (or *ab initio*) protein structure prediction on 3D Hydrophobic-Polar (HP) lattice model using evolutionary algorithms (Krasnogor, Hart, Smith, & Pelta, 1999). In other words, only sequence data is considered.

Different protein structure models have been proposed in the past (Silverio, 2008). Their differences mainly lie in their resolution levels and search space freedom. At the highest resolution level, all the atoms and bond angles can be simulated using molecular dynamics. Nevertheless, there is no free lunch. The simulation is hard to be completed by the current computational power. On the other hand, a study indicated that protein folding mechanisms might be simpler than previously thought (Baker, 2000). Thus this book chapter focuses on HP lattice model to capture the physical principles of protein folding process (Duan & Kollman, 2001).

In this problem, it assumes that the main driving forces are the interactions among the hydrophobic amino acid residues. The twenty amino acids are experimentally classified as either hydrophobic (H) or polar (P). An amino acid sequence is thus represented as a string {H,P}$^+$. Each residue is represented as a non-overlapping bead in a cubic lattice L. Each peptide bond in the main chain is represented as a connecting line. A protein is thus represented as a non-overlapping chain in L.

Based on the above model, the objective of the protein structure prediction problem is to find the conformation with the minimal energy for each protein. Mathematically, it is to minimize the following function (Li, Helling, Tang, & Wingreen, 1996):

$$H = \sum_{i+1<j} E(r_i, r_j)\, \Delta(r_i, r_j)$$

where $r_i$ and $r_j$ are amino acid residues at sequence position $i$ and $j$. The constraint $i+1<j$ is to ensure that $r_i$ and $r_j$ are not next to each other on their sequence and they are examined together once only. $\Delta(r_i, r_j) = 1$ when $r_i$ and $r_j$ are adjacent in L; Otherwise $\Delta(r_i, r_j) = 0$. As stated in the previous section, each residue is represented as either H or P. Thus $E(\sigma_i, \sigma_j)$ could be $E(H,H)$, $E(H,P)$, $E(P,H)$, or $E(P,P)$. For their values, three schemes have been proposed. The most widely used scheme is $E(H,H) = -1$, $E(H,P) = 0$, $E(P,H) = 0$, and $E(P,P) = 0$. The second scheme $E(H,H) = -2.3$, $E(H,P) = -1$, $E(P,H) = -1$, and $E(P,P) = 0$ was proposed. The last scheme $E(H,H) = -2$, $E(H,P) = 1$, $E(P,H) = 1$, and $E(P,P) = 1$ is called functional model protein (or ``shifted'' HP model) (Cutello, Nicosia, Pavone, & Timmis, 2007). As mentioned in (Silverio, 2008), the results are insensitive to the value of $E(H,H)$ as long as the physical constraints (Li, Helling, Tang, & Wingreen, 1996) are satisfied. Thus we have chosen the first scheme in this book chapter.

For the representation of an amino acid residue sequence, there are two conditions to be satisfied: (Krasnogor, Hart, Smith, & Pelta, 1999) (1) Sequence connectivity (2) Self-avoidance. Among the proposed representations (Cutello, Nicosia, Pavone, & Timmis, 2007), *Internal Coordinate* should be a favorable choice since it can handle the first condition implicitly. Internal coordinate is a representation system which residue positions depend on their sequence-predecessor residues. There are two types of internal coordinate representation: *Absolute Encoding* and *Relative Encoding*. Absolute encoding represents each residue position as the absolute direction from the previous residue. A sequence is represented as $\{U,D,L,R,F,B\}^{n-1}$ (Up, Down, Left, Right, Forward, Backward) (Unger & Moult, 1993). On the other hand, relative encoding represents those as relatively directional changes based on the directions of the two predecessor residues. Backward direction is omitted for one-step self-avoiding. Thus a sequence is represented as $\{F,R,L,U,D\}^{n-2}$ (Patton, Punch III, & Goodman, 1995). Except the forward move, a cyclic conformation is formed if a move is repeated four times. Krasnogor et al. (Krasnogor, Hart, Smith, & Pelta, 1999) have examined both representations on square lattices. Their results showed that relative encoding had better performance than absolute encoding on square lattices.

Although the 3D HP model seems relatively simple among other models, it has been proved that the protein structure prediction problem on the model is NP-Complete (Berger & Leighton, 1998). Thus researchers propose heuristics as compromising solutions. In particular, the seminal work by Unger et al. (Unger & Moult, 1993) experimentally showed that genetic algorithm approaches were better than Monte Carlos simulations. Thus researchers tried genetic algorithm as one of the heuristics to solve the problem. Nevertheless, the genetic algorithm approach by Unger et al. (Unger & Moult, 1993) was actually hybridized with Monte Carlo moves. Hence Patton et al. (Patton, Punch III, & Goodman, 1995) further generalized it into a standard genetic algorithm approach, which search space included infeasible regions penalized by a penalty function. Furthermore, they proposed ``relative encoding'' so that one-step self-avoiding constraints could be implicitly incorporated in the genome representation. Few years later, Krasnogor et al. (Krasnogor, Hart, Smith, & Pelta, 1999) published a work discussing the basic algorithmic factors affecting the problem. Since then, researchers explored different ways to tackle the problem. For instance, Krasnogor et al. further applied a multimeme algorithm, which adaptively chose multiple local searchers to reach optimal structures (Krasnogor, Blackburnem, Hirst, & Burke, 2002). Cox et al. (Cox, Mortimer-Jones, Taylor, & Johnston, 2004) and Hoque et al. (Hoque, Chetty, & Dooley, 2006) utilized heavy machinery of specific genetic operators and techniques. Ant colony algorithm (Shmygelska & Hoos, 2005), differential evolution (Bitello & Lopes, 2006), immune algorithm (Cutello, Nicosia, Pavone, & Timmis, 2007) and estimation of distribution algorithm (Santana, Larranaga, & Lozano, 2008) were also customized and reported in literatures. In particular, diversity preserving techniques were often incorporated in them. For instance, Duplicate Predator (Cox, Mortimer-Jones, Taylor, & Johnston, 2004), Aging Operator (Cutello, Nicosia, Pavone, & Timmis, 2007), and additional renormalization of the pheromone (Shmygelska & Hoos, 2005). They can be deemed as the signs of the multimodality in the problem. However, the necessity of multimodal optimization techniques has not been emphasized.

For the protein structure prediction problem, it is generally believed that the native state of protein should be at the conformation with the lowest energy. Thus previous works mainly focus on the minimal energy they could achieve: the minimal energy ever found and the average and standard deviation of the minimal energy across several runs. Nevertheless, Jahn et al. (Jahn & Radford, 2008) has shown that the native state is not necessarily a single global optimum. It may also be a local optimum in Fig.1 of (Jahn & Radford, 2008). For the HP lattice model, Unger et al. (Unger & Moult, 1993) have observed that there can be multiple conformations for each energy value. A recent fitness landscape study also indicated that HP landscapes were highly multimodal (Flores & Smith, 2003). Thus Wong et al. have proposed multimodal optimization techniques for the protein structure prediction problem (Wong, Leung, & Wong, 2010).

The most widely used distance measure should be the root mean square deviation (RMSD) (Holm & Sander, 1993). RMSD calculates the average absolute distances between two superimposed conformations' points. Nevertheless, if two conformations differ by only one point direction in relative encoding, their RMSD cannot reflect such small change. For instance, some conformations of the benchmark UM20 (Cotta, 2003) are visualized in Figure 4.

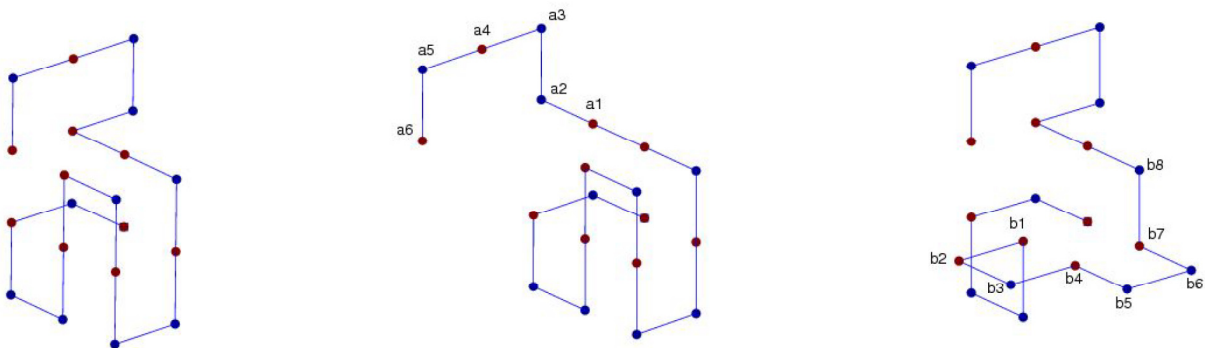

*Figure 4. The left most confirmation depicts an optimal conformation (LDLDFLUFDDFRFRDDFD) on the benchmark UM20. The other two confirmations (LDLDFLUFDDFRFEDDFD and LDLDDLLRLLDRFRDDFD) depict two candidate conformations after mutations. Red beads denote hydrophobic residues (H) while blue beads denote polar residues (P).*

To be mutated to the optimal conformation, Example A is only needed to change its move between a1 and a2 to R whereas example b is needed to change nearly all of its moves between b1 and b8. However, the RMSD of Example A with the optimal conformation (5 diagonal point changes a2 to a6) is larger than that of example B (4 diagonal point changes b2,b3,b5,b6). RMSD cannot capture the move information in relative encoding. Furthermore, if RMSD is applied, it will be quite computationally intensive. To calculate the RMSD between two conformations, the corresponding relative encoding genomes are converted to absolute 3D coordinates. Once converted, one of them is then translated and rotated to be optimally superimposed on the other. RMSD is then calculated which involves multiplications and square root calculations. In contrast, Hamming distance calculates the move differences between two relative encoding genomes. It is relatively computational tractable. Thus Hamming distance is usually adopted in this problem.

Basically there are two approaches for handling infeasible conformations: (1) Delete infeasible conformations (2) Tolerate infeasible conformations by adjusting their energy values by a penalty score (either constant or adaptive). Both approaches were thought beneficial in different view angles (Flores & Smith, 2003). For the first approach, it is conjectured that search space can be narrowed if infeasible conformations are deleted. For the second approach, it is conjectured that the paths to optimal conformations are shorter if infeasible conformations exist. Nevertheless, the study in (Flores & Smith, 2003) had a detailed analysis supporting the first approach.

## Evolutionary Algorithms for Protein-DNA Pattern Discovery

Protein-DNA interactions are essential in genetic activities such as transcription, packaging, rearrangement, and replication (Luscombe & Thornton, 2002). Understanding them forms the basis for further deciphering biological systems. In particular, the protein-DNA interactions between Transcription Factors (TFs) and Transcription Factor Binding Sites (TFBSs) play a central role in gene transcription. TFs bind in a sequence-specific manner to TFBSs to regulate gene transcription (Luscombe, Austin, Berman, & Thornton, 2000).

Nevertheless, it is expensive and laborious to experimentally identify the TF-TFBS binding sequence pairs, for example, using DNA footprinting (Galas & Schmitz, 1987) or gel electrophoresis (Garner & Revzin, 1981). The technology of Chromatin immunoprecipitation (ChIP) (Smith, Sumazin, Das, & Zhang, 2005) measures the binding of a particular TF to DNA of co-regulated genes on a genome-wide scale *in vivo*, but at low resolution. Further processing is needed to extract precise TFBSs (Liu, Brutlag, & Liu, 2002). To share the precious sequence data, researchers have built databases. In particular, TRANSFAC (Matys V. , et al., 2006) is one of the largest and most representative databases for regulatory elements including TFs, TFBSs, nucleotide distribution matrices of the TFBSs, and regulated genes. The data are expertly annotated and manually corrected from peer-reviewed publications and experimentally verified studies. Other annotation databases of TF families and binding domains are also available (e.g. PROSITE (Hulo, et al., 2008), Pfam (Bateman, et al., 2004)).

On the other hand, high-quality TF-TFBS binding structures can provide valuable insights into putative principles of binding. However, it is difficult and time consuming to extract those high-resolution 3D TF-TFBS complex structures with X-ray crystallography (Smyth & Martin, 2000) or Nuclear Magnetic Resonance (NMR) spectroscopic analysis (Mohan & Hosur, 2009). To share the precious structural data, researchers have also built databases. In particular, the Protein Data Bank (PDB) (Berman, et al., 2000) serves as a representative repository of such experimentally extracted protein-DNA (in particular TF-TFBS) complexes with high resolution at atomic levels. However, the available 3D structures are far from complete. As a result, there is strong motivation to have automatic methods, particularly, computational approaches based on existing abundant data, to provide testable candidates of TF-TFBS binding sequence pairs with high confidence to guide and accelerate the wet-lab experiments.

Most of the previous computational attempts related to TF-TFBS interactions are devoted to discover either the motifs of TF domains or those of TFBSs separately. The TF domains and TFBSs sequences are somewhat conserved due to their functional similarity and importance. By exploiting the conservation, computational methods called motif discovery have been proposed to save the expensive and laborious laboratory experiments (MacIsaac & Fraenkel, 2006). The methods usually make use of comprehensive statistical and scoring models to extract the domain information from the background sequences (Jensen S. T., Liu, Zhou, & Liu, 2004). In addition, data mining methods have been proposed to find the sequence pairs. For instance, support vector machines (SVM) (Ofran, Mysore, & Rost, 2007) and regressions (Zhou & Liu, 2008). Distinct from motif discovery, they utilize the biochemical information in sequence data (e.g. base compositions, structures, thermodynamic properties (Ahmad, Keskin, Sarai, & Nussinov, 2008)) to perform prediction. Nevertheless, most of their results are not concrete sequences (the most explicit and interpretable format).

Thus Leung et al. have proposed a framework based on association rule mining with Apriori algorithm (Agrawal, Imielinski, & Swami, 1993) to discover associated TF-TFBS binding sequence patterns in the most explicit and interpretable form from TRANSFAC (Leung, et al., 2010). With downward closure property, the algorithm guarantees the exact and optimal performance to generate all frequent TFBS k-mer TF k-mer pairs from TRANSFAC where a k-mer is a string with length equal to k. The approach relies merely on sequence information without any prior knowledge in TF binding domains or protein-DNA 3D structure data. From comprehensive evaluations, statistics of the discovered patterns are shown to reflect meaningful binding characteristics. According to independent literature, PDB data and homology modeling, a good number of TF-TFBS binding patterns discovered have been verified by experiments and annotations. They exhibit atomic-level interactions between the respective TF binding domains and specific nucleotides of the TFBS from experimentally determined protein-DNA 3D structures.

Although the above, the sequence pairs discovered are in one-to-one mappings (Leung, et al., 2010). One TF amino acid sequence is coupled with one TFBS DNA sequence. In the biological world, a TF may bind to a promoter using several contact surface subsequences. Some surfaces of the TF may also be interacting surfaces to recruit another TF as a performing complex (White R. J., 2001). For instance, McGuire et al found that there were two conserved parts for the ArcA-P recognition motif in E.coli (McGuire, De Wulf, Church, & Lin, 1999). Kato et al. proposed a novel method to identify combinatorial regulation of transcription factors and binding motifs using chromatin immunoprecipitation (ChIP) data with microarray expression data (Kato, Hata, Banerjee, Futcher, & Zhang, 2004). A case study in the evolution of combinatorial gene regulation in Fungi has also been carried out (Tuch, Galgoczy, Hernday, Li, & Johnson, 2008). Biochemists have used biochemical experiment methods to observe many evidences. For instance, SOX proteins perform their function in a complex interplay with other transcription factors in a manner highly dependent on cell type and promoter context. In particular, multiple TFBSs are found within the enhancer of the FGF4 gene during early embryonic expression. One is a recognition element for POU proteins; the other is a binding site for SOX proteins. The POU and SOX protein partnership is crucial to determine cell fate. Scientists have also used 3D structural determination methods to observe such combinatorial behavior. Some examples can be found in (Kato, Hata, Banerjee, Futcher, & Zhang, 2004). Many experimental evidences can also be found in TransCompel (Matys V. , et al., 2006) which is a comprehensive database on the composite interactions between TFs binding to their TFBSs. Multiple TF amino acid sequences may be coupled with multiple TFBS DNA sequence, instead of just one-to-one mapping. Considering the huge search space, Wong et al. have further proposed an evolutionary algorithm to learn generalized representations from the original pairs (Wong, Peng, Wong, & Leung, 2011). In particular, the original pairs are evolved to pairs of boolean expressions (trees) of k-mers An example is shown in Figure 5.

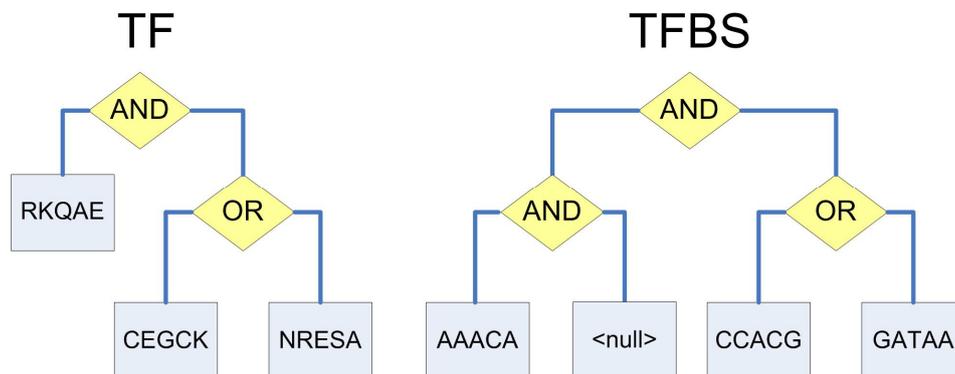

*Figure 5. Exemplary pair of boolean expressions (trees) of k-mers. The left tree is on the TF side with amino acid k-mers while the right tree is on the TFBS side with DNA k-mers.*

Evolving trees (e.g. boolean expressions of k-mers) by evolutionary algorithms are well studied in the genetic programming field. Many design issues have been reviewed in section 2 of (Banzhaf, Nordin, Keller, & Francone, 1998). In particular, researchers are especially concerned about the roles of crossovers and mutations. Some of them argue that crossovers are not beneficial to the evolution, whereas the others hold the opposite view (Spears & Anand, 1991). Some of them also argue that mutations are not needed, whereas the others hold the opposite view (Poli, Langdon, & Mcphee, A Field Guide to Genetic Programming, 2008). Even extensive experiments on comparing crossovers and mutations on a series of well known problems have been conducted (Luke & Spector, 1998). The results can only reflect that it is problem-dependent (White & Poulding, 2009). The debate is still continuing. Thus, as a compromising solution, both crossover and mutation operators are adopted in (Wong, Peng, Wong, & Leung, 2011). Another important topic in genetic programming is to control the ``bloat'' property. During a typical run of genetic programming, it is often found that some unnecessary components (called ``introns'') are formed. It is intuitive for us to think that they are not necessary, and thus not good. Soule et al. (Terence, Foster, & Dickinson, 1996) suggested a fitness function which penalizes trees with many introns. Rosca also suggested parsimony pressure on selecting trees was beneficial to grow toward the optimal structures (Rosca, 1997). Some researchers also suggest that the presence of the introns can confuse crossover operators, protecting good modules/components as discussed in Chapter 7 of (Banzhaf, Nordin, Keller, & Francone, 1998). For instance, some of them made use of the bloat property to evolve buffer overflow attack codes which can successfully hide themselves from intrusion detectors (Kayacik, Heywood, & Zincir-Heywood, 2006). Indeed, the bloat property is a double-edged sword.

## CONCLUSION

Evolutionary algorithms build a bridge between computer science and nature (Kari & Rozenberg, 2008). Instead of artificial creation, evolutionary algorithm emphasizes on learning from nature. Nature rules are applied or modeled to build novel computational techniques, which can be well adapted and integrated into different contexts. For instance, inspired from the Darwin's evolutionary theory, John Holland has proposed genetic algorithm which simulates the evolutionary process for natural selection. It has been proved successful in different applications widely. Indeed, the design of evolutionary algorithms draws inspiration from nature. They resemble the natural mechanism and are affiliated to nature. It is not surprising for us to expect that they should be among the best methods in bioinformatics to decipher nature in the future.

Besides, we suspect that most of the evolutionary algorithms applied to bioinformatics are always stuck in local optima. People are either not aware of the issue, or too lazy to study and handle it as long as the local optima found are good enough in practice. Thus we can foresee, if we further apply some evolutionary algorithms for multimodal optimization to the bioinformatics problems, promising results will be probably obtained.

Notably, Wong et al. have proposed robust and competitive methods for multimodel optimization (Wong, Wu, Mok, Peng, & Zhang, 2012). In those methods, to explore the locality principle in evolutionary computation, crowding differential evolution (CrowdingDE) is incorporated with locality for multimodal optimization. Instead of generating trial vectors randomly, the first method proposed takes advantage of spatial locality to generate trial vectors. Temporal locality is also adopted to help generate offspring in the second method proposed. Temporal and spatial localities are then applied together in the third method proposed.

Here, we briefly describe those methods one by one. If diversity maintenance techniques are not applied, most evolutionary algorithms will prematurely converge and get stuck in a local optimum. To cope with the problem, the algorithms are usually equipped with their own local operations for diversity maintenance. In CorwdingDE, the local operation is the crowding technique, in which each offspring can only replace the individual which is most similar to itself. Looking at this technique more deeply, a restriction is proposed in the individual replacement method so that a individual gets replaced only when another individual is generated and evaluated fitter than the former within the same niche. The algorithm is forced to passively wait for the trial vector generations for feasible replacements. Unfortunately, the trial vector generations of CrowdingDE are random, as stated before. Thus the fundamental computer science concept, spatial locality, is applied to the trial vector generations in order to increase the chance of successful replacements in this work.

Close individuals tend to have similar characteristics. In a run of evolutionary algorithm, the population is usually divided into different niches. Within each niche, the individuals exhibit similar positions and step-sizes for improvement. After several generations, the differences between niches may be large. It will be a disaster if a single evolutionary strategy is applied to all of them regardless of their niches. Interestingly, it is a double-edged sword. Such a property also gives us spatial locality: the crossovers between close individuals can have higher chance to generate better offspring in their niche, comparing with the crossovers between distinct individuals. Thus the individuals which are closer to the parent than the others in the same generation should be given more chance of trial vector generations within a population. We should be aware that such a evolutionary policy may not be applicable for the problem domains other than multimodal optimization because the selected individuals may be similar to each other. Such a similarity may generally reduce the step size. To bring such a neighborhood idea into trial vector generations, spatial locality is proposed as a measure for selecting individuals to form trial vectors. The distances between the parent and all candidate individuals are computed and transformed into proportions which form a roulette-wheel. Within the roulette-wheel, a larger portion is given to the candidate individual which is closer to the parent than the others in the same generation. It follows that closer individuals are given higher chance of trial vector generations and vice versa.

By doing so, each trial vector generation becomes a local operation tailor-made for the parent individual. Crowding Differential Evolution (CrowdingDE) is reformulated as a hybrid algorithm which takes advantage of spatial locality. Thus we call the proposed algorithm as Crowding Differential Evolution using Spatial Locality (CrowdingDE-SL).

Besides spatial locality, temporal locality is also an intrinsic feature we can make use of. For instance, the most typical application is the use of cache in a computer system. If some data is accessed at a given time, then it is very likely that these data will be referenced again. Thus it is useful to store these data into high speed caches. Based on the same idea, such a temporal locality concept can also be incorporated into CrowdingDE. If an individual is replaced by another fitter individual under the crowding selection, then the vector difference between them is a improvement step within their niche. If the step is reused in a correct situation, it can contribute to improvements within their niche again. Thus it is advantageous for an algorithm to save and reuse these vectors for improvement. Nevertheless, as several generations pass by, a vast amount of these vectors are accumulated. It is impossible to store them all. As an intuitive solution, these vectors should be summarized on the fly. To do that, there are lots of existing techniques available. Considering the heavily iterative property of evolutionary algorithms, computational efficiency needs be taken into account seriously. Thus a simple summation technique with a discount factor is proposed.

In that technique, each individual is allocated with an array called delta which is of the same size as the genome. The main use of the array is to store the temporal locality history. Whenever an offspring generated is fitter than its nearest neighbor, the method records and stores their vector difference, plus the array delta of the nearest neighbor (with a discount factor), into the array delta of the offspring. After that, one more offspring is generated by summing the genome of the offspring and the array delta of the offspring together. If the new offspring is fitter than the original offspring, then the new offspring replaces the nearest neighbor. Otherwise, the original offspring replaces the nearest neighbor. Combined with this local operation, Crowding Differential Evolution (CrowdingDE) is reformulated as a hybrid algorithm which takes advantage of temporal locality. Thus we call it Crowding Differential Evolution using Temporal Locality (CrowdingDE-TL).

Having incorporated spatial and temporal locality into CrowdingDE separately, it is intuitive for us to apply them together since they belong to different modules: spatial locality takes effect in trial vector generations, whereas temporal locality takes effect in the selection stage after trial vector generations. Thus CrowdingDE can be combined with both spatial and temporal locality together, which is subsequently called Crowding Differential Evolution using Spatial and Temporal Locality (CrowdingDE-STL).

Numerical experiments are conducted to compare the proposed methods with the state-of-the-art methods on benchmark functions extensively. Experimental analysis is undertaken to observe the effect of locality and the synergy between temporal locality and spatial locality. Further experiments are also conducted on two application problems. One is the varied-line-spacing holographic grating design problem, while the other is the protein structure prediction problem. The numerical results demonstrate the effectiveness of those methods (Wong, Wu, Mok, Peng, & Zhang, 2012).

## REFERENCES

Abecasis, G. R., Auton, A., Brooks, L. D., DePristo, M. A., & Durbin, R. M. (2012). An integrated map of genetic variation from 1,092 human genomes. *Nature , 491* (7422), 56-65.
Abeel, T., Van de Peer, Y., & Saeys, Y. (2009). Toward a gold standard for promoter prediction evaluation. *Bioinformatics , 25* (12), i313--i320.
Agrawal, R., Imielinski, T., & Swami, A. (1993). Mining association rules between sets of items in large databases. *SIGMOD '93: Proceedings of the 1993 ACM SIGMOD international conference on Management of data ,* 207-216.
Ahmad, S., Keskin, O., Sarai, A., & Nussinov, R. (2008). Protein-DNA interactions: structural, thermodynamic and clustering patterns of conserved residues in DNA-binding proteins. *Nucleic Acids Res. , 36*, 5922-5932.
Altschul, S. F., Gish, W., Miller, W., Myers, E. W., & Lipman, D. J. (1990). Basic local alignment search tool. *Journal of molecular biology , 215* (3), 403-410.
Aluru, S. (2005). *Handbook of Computational Molecular Biology (Chapman \& All/Crc Computer and Information Science Series).* Chapman \& Hall/CRC.
Anfinsen, C. B. (1973). Principles that Govern the Folding of Protein Chains. *Science , 181* (4096), 223-230.
Baker, D. (2000). A surprising simplicity to protein folding. *Nature , 405* (6782), 39-42.
Baker, D., & Sali, A. (2001). Protein Structure Prediction and Structural Genomics. *Science , 294* (5540), 93-96.
Banzhaf, W., Nordin, P., Keller, R. E., & Francone, F. D. (1998). *Genetic Programming -- An Introduction; On the Automatic Evolution of Computer Programs and its Applications.* San Francisco, CA, USA: Morgan Kaufmann.

Bateman, A., Coin, L., Durbin, R., Finn, R. D., Hollich, V., GrifRths-Jones, S., et al. (2004). The Pfam protein families database. *Nucleic Acids Res , 32*, D138--141.

Berger, B., & Leighton, T. (1998). Protein folding in the hydrophobic-hydrophilic (HP) is NP-complete. *RECOMB '98: Proceedings of the second annual international conference on Computational molecular biology* (pp. 30-39). New York, NY, USA: ACM.

Berman, H. M., Westbrook, J., Feng, Z., Gilliland, G., Bhat, T. N., Weissig, H., et al. (2000). The Protein Data Bank. *Nucl. Acids Res. , 28* (1), 235-242.

Bertini, I., De, M., Moretti, F., & Pizzuti, S. (2010). Start-Up Optimisation of a Combined Cycle Power Plant with Multiobjective Evolutionary Algorithms. *EvoApplications (2)*, (pp. 151-160).

Beyer, H., & Schwefel, H. (2002). Evolution strategies - A comprehensive introduction. *Natural Computing , 1* (1), 3-52.

Bitello, R., & Lopes, H. S. (2006). A Differential Evolution Approach for Protein Folding. *Computational Intelligence and Bioinformatics and Computational Biology, 2006. CIBCB '06. 2006 IEEE Symposium on*, (pp. 1-5). Toronto, Ont.,.

Civicioglu, P., & Besdok, E. (2013). A conceptual comparison of the Cuckoo-search, particle swarm optimization, differential evolution and artificial bee colony algorithms. *Artificial Intelligence Review , 39* (4), 315-346.

Consortium, E. (2012). An integrated encyclopedia of DNA elements in the human genome. *Nature , 489* (7414), 57-74.

Consortium, I. H. (2004). Finishing the euchromatic sequence of the human genome. *Nature , 431* (7011), 931-945.

Cotta, C. (2003). Protein Structure Prediction Using Evolutionary Algorithms Hybridized with Backtracking. *IWANN '03: Proceedings of the 7th International Work-Conference on Artificial and Natural Neural Networks* (pp. 321-328). Berlin, Heidelberg: Springer-Verlag.

Cox, G. A., Mortimer-Jones, T. V., Taylor, R. P., & Johnston, R. L. (2004). Development and optimisation of a novel genetic algorithm for studying model protein folding. *Theoretical Chemistry Accounts: Theory, Computation, and Modeling , 112* (3), 163-178.

Cutello, V., Nicosia, G., Pavone, M., & Timmis, J. (2007). An Immune Algorithm for Protein Structure Prediction on Lattice Models. *{IEEE} Transactions on Evolutionary Computation , 11* (1), 101-117.

De Jong, K. A. (2006). *Evolutionary Computation. A Unified Approach.* Cambridge, MA, USA: MIT Press.

Del Venayagamoorthy, G. K., Mohagheghi, S., Hernandez, J. C., & Harley, R. G. (2008). Particle Swarm Optimization: Basic Concepts, Variants and Applications in Power Systems. *Evolutionary Computation, IEEE Transactions on , 12* (2), 171-195.

D'Haeseleer, P., Liang, S., & Somogyi, R. (2000). Genetic network inference: from co-expression clustering to reverse engineering. *Bioinformatics (Oxford, England) , 16* (8), 707-726.

Dorigo, M., & Gambardella, L. M. (1997). Ant colony system: a cooperative learning approach to the traveling salesman problem. *Evolutionary Computation, IEEE Transactions on , 1* (1), 53-66.

Duan, Y., & Kollman, P. A. (2001). Computational protein folding: from lattice to all-atom. *IBM Syst. J. , 40* (2), 297-309.

Esling, P., Carpentier, G., & Agon, C. (2010). Dynamic Musical Orchestration Using Genetic Algorithms and a Spectro-Temporal Description of Musical Instruments. *EvoApplications (2)*, (pp. 371-380).

Feoktistov, V. (2006). *Differential Evolution: In Search of Solutions (Springer Optimization and Its Applications)*. Secaucus, NJ, USA: Springer-Verlag New York, Inc.

Fleischmann, R. D., Adams, M. D., White, O., Clayton, R. A., Kirkness, E. F., Kerlavage, A. R., et al. (1995). {{W}hole-genome random sequencing and assembly of {H}aemophilus influenzae {R}d}. *Science , 269*, 496-512.

Flores, S. D., & Smith, J. (2003). Study of fitness landscapes for the HP model of protein structure prediction. *Evolutionary Computation, 2003. {CEC} '03. The 2003 Congress on, 4*, pp. 2338-2345.

Furuholmen, M., Glette, K., Hovin, M., & Torresen, J. (2010). Evolutionary Approaches to the Three-dimensional Multi-pipe Routing Problem: A Comparative Study Using Direct Encodings. *EvoCOP*, (pp. 71-82).

Galas, D. J., & Schmitz, A. (1987). DNAse footprinting: a simple method for the detection of protein-{DNA} binding specificity. *Nucleic Acids Res. , 5* (9), 3157-3170.

Garner, M. M., & Revzin, A. (1981). A gel electrophoresis method for quantifying the binding of proteins to specific {DNA} regions: application to components of the Escherichia coli lactose operon regulatory system. *Nucleic Acids Res. , 9* (13), 3047-3060.

Goffeau, A., Barrell, B., Bussey, H., Davis, R., Dujon, B., Feldmann, H., et al. (1996). {{L}ife with 6000 genes}. *Science , 274*, 563-567.

Goldberg, D. E., & Richardson, J. (1987). Genetic algorithms with sharing for multimodal function optimization. *Proceedings of the Second International Conference on Genetic algorithms and their application* (pp. 41-49). Hillsdale, NJ, USA: L. Erlbaum Associates Inc.

GuhaThakurta, D. (2006). Computational identification of transcriptional regulatory elements in DNA sequence. *Nucleic Acids Res. , 34*, 3585-3598.

Hadka, D., & Reed, P. (2013). Borg: An Auto-Adaptive Many-Objective Evolutionary Computing Framework. *Evolutionary Computation , 21* (2), 231-259.

Holland, J. H. (1975). *Adaptation in natural and artificial systems.* Ann Arbor: University of Michigan Press.

Holm, L., & Sander, C. (1993). Protein structure comparison by alignment of distance matrices. *J. Mol. Biol. , 233*, 123-138.

Hoque, T., Chetty, M., & Dooley, L. S. (2006). A Guided Genetic Algorithm for Protein Folding Prediction Using 3D Hydrophobic-Hydrophilic Model. *Evolutionary Computation, 2006. CEC 2006. IEEE Congress on*, (pp. 2339-2346). Vancouver, BC,.

Hulo, N., Bairoch, A., Bulliard, V., Cerutti, L., Cuche, B. A., de Castro, E., et al. (2008). The 20 years of PROSITE. *Nucl. Acids Res. , 36* (suppl\_1), D245--249.

Initiative, A. G. (2000). Analysis of the genome sequence of the flowering plant Arabidopsis thaliana. *Nature , 408*, 796-815.

Iredi, S., Merkle, D., & Middendorf, M. (2000). Bi-Criterion Optimization with Multi Colony Ant Algorithms. *in Proceedings of the First International Conference on Evolutionary Multi-Criterion Optimization (EMO 2001), no. 1993 in LNCS* (pp. 359-372). Springer.

Jahn, T. R., & Radford, S. E. (2008). Folding versus aggregation: polypeptide conformations on competing pathways. *Arch. Biochem. Biophys. , 469*, 100-117.

Jensen, S. T., Liu, X. S., Zhou, Q., & Liu, J. S. (2004). Computational discovery of gene regulatory binding motifs: a Bayesian perspective. *Statistical Science , 19* (1), 188-204.

Karaboga, D., Akay, B., & Ozturk, C. (2007). Artificial Bee Colony (ABC) Optimization Algorithm for Training Feed-Forward Neural Networks. In V. Torra, Y. Narukawa, & Y. Yoshida (Ed.), *MDAI. 4617*, pp. 318-329. Springer.

Kari, L., & Rozenberg, G. (2008). The many facets of natural computing. *Commun. ACM , 51* (10), 72-83.

Kato, M., Hata, N., Banerjee, N., Futcher, B., & Zhang, M. Q. (2004). Identifying combinatorial regulation of transcription factors and binding motifs. *Genome Biol. , 5*, R56.

Kayacik, H., Heywood, M., & Zincir-Heywood, N. (2006). On evolving buffer overflow attacks using genetic programming. *Proceedings of the 8th annual conference on Genetic and evolutionary computation* (pp. 1667-1674). New York: ACM.

Krasnogor, N., Blackburnem, B., Hirst, J., & Burke, E. (2002). Multimeme Algorithms for Protein Structure Prediction. *7th International Conference Parallel Problem Solving from Nature. 2439*, pp. 769-778. Granada, Spain: Springer Berlin / Heidelberg.

Krasnogor, N., Hart, W., Smith, J., & Pelta, D. (1999). Protein Structure Prediction With Evolutionary Algorithms. *International Genetic and Evolutionary Computation Conference (GECCO99)* (pp. 1569-1601). Morgan Kaufmann.

Lampinen, J., & Zelinka, I. (1999). Mechanical engineering design optimization by differential evolution. *New ideas in optimization* , 127-146.

Larkin, F., & Ryan, C. (2010). Modesty Is the Best Policy: Automatic Discovery of Viable Forecasting Goals in Financial Data. *EvoApplications (2)*, (pp. 202-211).

Leung, K. S., Wong, K. C., Chan, T. M., Wong, M. H., Lee, K. H., Lau, C. K., et al. (2010). Discovering protein-DNA binding sequence patterns using association rule mining. *Nucleic Acids Res. , 38* (19), 6324-6337.

Levinthal, C. (1968). Are there pathways for protein folding? *J. Chem. Phys. , 65*, 44-45.

Li, H., Helling, R., Tang, C., & Wingreen, N. (1996). Emergence of Preferred Structures in a Simple Model of Protein Folding. *Science , 273* (5275), 666-669.

Liang, Y., Leung, K. S., & Mok, T. S. (2008). Evolutionary drug scheduling models with different toxicity metabolism in cancer chemotherapy. *Appl. Soft Comput. , 8* (1), 140-149.

Liu, X. S., Brutlag, D. L., & Liu, J. S. (2002). An algorithm for finding protein--{DNA} binding sites with applications to chromatinimmunoprecipitation microarray experiments. *Nat. Biotechnol. , 20*, 835-839.

Luke, S., & Spector, L. (1998). A Revised Comparison of Crossover and Mutation in Genetic Programming. *Proceedings of the Third Annual Genetic Programming Conference (GP98)* (pp. 208-213). Morgan Kaufmann.

Luscombe, N. M., & Thornton, J. M. (2002). Protein-DNA interactions: amino acid conservation and the effects of mutations on binding specificity. *J Mol Biol , 320* (5), 991-1009.

Luscombe, N. M., Austin, S. E., Berman, H. M., & Thornton, J. M. (2000). An overview of the structures of protein-DNA complexes. *Genome Biol. , 1* (1), REVIEWS001.

MacIsaac, K. D., & Fraenkel, E. (2006). Practical strategies for discovering regulatory DNA sequence motifs. *PLoS Comput Biol. , 2* (4), e36.

Matys, V., Kel-Margoulis, O. V., Fricke, E., Liebich, I., Land, S., Barre-Dirrie, A., et al. (2006). TRANSFAC and its module TRANSCompel: transcriptional gene regulation in eukaryotes. *Nucleic Acids Research , 34*, 108-110.

McGuffin, L. J., Bryson, K., & Jones, D. T. (2000). The PSIPRED protein structure prediction server. *Bioinformatics (Oxford, England) , 16* (4), 404-405.

McGuire, A. M., De Wulf, P., Church, G. M., & Lin, E. C. (1999). A weight matrix for binding recognition by the redox-response regulator ArcA-P of Escherichia coli. *Mol. Microbiol. , 32*, 219-221.

Mohan, P. M., & Hosur, R. V. (2009). Structure-function-folding relationships and native energy landscape of dynein light chain protein: nuclear magnetic resonance insights. *J. Biosci.*, *34*, 465-479.

Niu, B., & Wang, H. (2012). Bacterial Colony Optimization. *Discrete Dynamics in Nature and Society*, 698057.

Ofran, Y., Mysore, V., & Rost, B. (2007). Prediction of DNA-binding residues from sequence. *Bioinformatics*, *23* (13), i347--353.

Patton, A. L., Punch III, W. F., & Goodman, E. D. (1995). A Standard GA Approach to Native Protein Conformation Prediction. *Proceedings of the 6th International Conference on Genetic Algorithms* (pp. 574-581). San Francisco, CA, USA: Morgan Kaufmann Publishers Inc.

Poli, R., Kennedy, J., & Blackwell, T. (2007). Particle swarm optimization. *Swarm Intelligence*, *1* (1), 33-57.

Poli, R., Langdon, W. B., & Mcphee, N. F. (2008). *A Field Guide to Genetic Programming.* Lulu Enterprises, UK Ltd.

Price, K. V. (1997). Differential evolution vs. the functions of the 2nd ICEO. *Evolutionary Computation, 1997., IEEE International Conference on*, (pp. 153-157). Indianapolis, IN, USA.

Rahmat-Samii, Y., & Michielssen, E. (Eds.). (1999). *Electromagnetic Optimization by Genetic Algorithms.* New York, NY, USA: John Wiley \& Sons, Inc.

Ronquist, F., & Huelsenbeck, J. P. (2003). MrBayes 3: Bayesian phylogenetic inference under mixed models. *Bioinformatics*, *19* (12), 1572-1574.

Rosca, J. P. (1997). Analysis of Complexity Drift in Genetic Programming. *Genetic Programming 1997: Proceedings of the Second Annual Conference* (pp. 286-294). Morgan Kaufmann.

Sacco, W. F., Henderson, N., Rios-Coelho, A. C., Ali, M. M., & Pereira, C. M. (2009, June). Differential evolution algorithms applied to nuclear reactor core design. *Annals of Nuclear Energy* .

Santana, R., Larranaga, P., & Lozano, J. A. (2008). Protein Folding in Simplified Models With Estimation of Distribution Algorithms. *{IEEE} Transactions on Evolutionary Computation*, *12* (4), 418-438.

Schoonderwoerd, R., Bruten, J. L., Holland, O. E., & Rothkrantz, L. J. (1996). Ant-based load balancing in telecommunications networks. *Adapt. Behav.*, *5* (2), 169-207.

Shah-Hosseini, H. (2009). The intelligent water drops algorithm: a nature-inspired swarm-based optimization algorithm. *International Journal of Bio-Inspired Computation*, *1*, 71-79.

Shmygelska, A., & Hoos, H. (2005). An ant colony optimisation algorithm for the 2D and 3D hydrophobic polar protein folding problem. *BMC Bioinformatics*, *6* (1), 30.

Silverio, H. (2008). Evolutionary Algorithms for the Protein Folding Problem: A Review and Current Trends. *Computational Intelligence in Biomedicine and Bioinformatics*, 297-315.

Smith, A. D., Sumazin, P., Das, D., & Zhang, M. Q. (2005). Mining ChIP-chip data for transcription factor and cofactor binding sites. *Bioinformatics*, *Suppl 1* (20), i403--i412.

Smyth, M. S., & Martin, J. H. (2000). X ray crystallography. *Molecular pathology : MP*, *53* (1), 8-14.

Spears, W. M., & Anand, V. (1991). A Study of Crossover Operators in Genetic Programming. *ISMIS '91: Proceedings of the 6th International Symposium on Methodologies for Intelligent Systems* (pp. 409-418). London, UK: Springer-Verlag.

Storn, R., & Price, K. (1997). Differential Evolution - A Simple and Efficient Heuristic for global Optimization over Continuous Spaces. *Journal of Global Optimization*, *11* (4), 341-359.

Terence, S., Foster, J. A., & Dickinson, J. (1996). Code growth in genetic programming. *Proceedings of the 1st annual conference on genetic programming* (pp. 215-223). Cambridge, MA: MIT Press.

Tuch, B. B., Galgoczy, D. J., Hernday, A. D., Li, H., & Johnson, A. D. (2008). The evolution of combinatorial gene regulation in fungi. *PLoS Biol. , 6*, e38.

Unger, R., & Moult, J. (1993). Genetic Algorithm for 3D Protein Folding Simulations. *Proceedings of the 5th International Conference on Genetic Algorithms* (pp. 581-588). San Francisco, CA, USA: Morgan Kaufmann Publishers Inc.

Vuswabatgab, G. M., Afanasyer, V., Buldyrev, S. V., Murphy, E. J., Prince, P. A., & Stanley, H. E. (1996). Levy flight search patterns of wandering albatrosses. *Nature , 381*, 413-415.

White, D. R., & Poulding, S. (2009). A Rigorous Evaluation of Crossover and Mutation in Genetic Programming. *EuroGP '09: Proceedings of the 12th European Conference on Genetic Programming* (pp. 220-231). Berlin, Heidelberg: Springer-Verlag.

White, R. J. (2001). *Gene Transcription: Mechanisms and Control.* Wiley-Blackwell.

Whitley, D. (1994). A genetic algorithm tutorial. *Statistics and Computing , 4* (2), 65-85.

Wong, K. C., & Zhang, Z. (2014). SNPdryad: predicting deleterious non-synonymous human SNPs using only orthologous protein sequences. *Bioinformatics , 30* (8), 1112-1119.

Wong, K. C., Chan, T. M., Peng, C., Li, Y., & Zhang, Z. (2013). DNA motif elucidation using belief propagation. *Nucleic Acids Res. , 41* (16), e153.

Wong, K. C., Leung, K. S., & Wong, M. H. (2009). An evolutionary algorithm with species-specific explosion for multimodal optimization. *GECCO '09: Proceedings of the 11th Annual conference on Genetic and evolutionary computation* (pp. 923-930). New York, NY, USA: ACM.

Wong, K. C., Leung, K. S., & Wong, M. H. (2010). Effect of Spatial Locality on An Evolutionary Algorithm for Multimodal Optimization. *EvoApplications 2010, Part I, LNCS 6024.* Springer-Verlag.

Wong, K. C., Leung, K. S., & Wong, M. H. (2010). Protein Structure Prediction on a Lattice Model via Multimodal Optimization Techniques. *Proceedings of the 12th annual conference on Genetic and evolutionary computation* (pp. 155-162). Portland: ACM.

Wong, K. C., Li, Y., Peng, C., & Zhang, Z. (2015). SignalSpider: probabilistic pattern discovery on multiple normalized ChIP-Seq signal profiles. *Bioinformatics , 31* (1), 17-24.

Wong, K. C., Peng, C., Li, Y., & Chan, T. M. (2014). Herd Clustering: a synergistic data clustering approach using collective intelligence. *Applied Soft Computing , 23*, 61-75.

Wong, K. C., Peng, C., Wong, M. H., & Leung, K. S. (2011). Generalizing and learning protein-DNA binding sequence representations by an evolutionary algorithm. *Soft Comput. , 15* (8), 1631-1642.

Zhou, Q., & Liu, J. S. (2008). Extracting sequence features to predict protein-DNA interactions: a comparative study. *Nucl. Acids Res. , 36* (12), 4137-4148.